\documentclass[sigconf]{acmart}
\AtBeginDocument{%
  \providecommand\BibTeX{{%
    \normalfont B\kern-0.5em{\scshape i\kern-0.25em b}\kern-0.8em\TeX}}}

\usepackage{subcaption}
\usepackage{graphicx}
\usepackage{blindtext}
\setcopyright{acmcopyright}
\copyrightyear{2024}
\acmYear{2024}
\acmDOI{XXXXXXX.XXXXXXX}

\acmConference[KDD]{ACM SIGKDD Conference on Knowledge Discovery \& Data Mining}{August 25--29, 2024}{Barcelona, Spain}
\acmBooktitle{KDD '24: AACM SIGKDD Conference on Knowledge Discovery \& Data Mining,
 August 25--29, 2023, Barcelona, Spain} 
\acmPrice{15.00}
\acmISBN{978-1-4503-XXXX-X/18/06}

\begin{document}

%\title{A Deep Learning Approach for Imbalanced Tabular Data in Advertising}

\title{A Deep Learning Approach for Imbalanced Tabular Data in Advertiser Prospecting: A Case of Direct Mail Prospecting}

%\author{qwerty}
%\email{qwerty@postie.com}
%\email{webmaster@marysville-ohio.com}
%\affiliation{%
%  \institution{Institute for Clarity in Documentation}
%  \streetaddress{P.O. Box 1212}
%  \country{USA}

%\author{Lars Th{\o}rv{\"a}ld}
%\affiliation{%
%  \institution{The Th{\o}rv{\"a}ld Group}
%  \streetaddress{1 Th{\o}rv{\"a}ld Circle}
%\email{larst@affiliation.org}

\author{Sadegh Farhang, William Hayes, Nick Murphy, Jonathan Neddenriep and Nicholas Tyris}
\affiliation{%
  \institution{Postie}
  \country{USA}
}
\email{{sadegh.farhang, william.hayes, nmurphy, jonathan, ntyris}@postie.com}

%\affiliation{%
% \institution{Rajiv Gandhi University}

%\affiliation{%
%  \institution{Tsinghua University}
%  \streetaddress{30 Shuangqing Rd}

\renewcommand{\shortauthors}{Sadegh Farhang, et al.}

\begin{abstract}

%Acquiring new customers is integral for each company to grow their business and increase their foothold. Prospecting is the process of identifying and contacting potential customers, which can be done in different ways, ranging from online advertisement to direct mail. Previous research has shown that direct mail is still one of the most effective ways to acquire new customers, which is the focus of this paper.

Acquiring new customers is a vital process for growing businesses. Prospecting is the process of identifying and marketing to potential customers using methods ranging from online digital advertising, linear television, out of home, and direct mail. Despite the rapid growth in digital advertising (particularly social and search), research shows that direct mail remains one of the most effective ways to acquire new customers. However, there is a notable gap in the application of modern machine learning techniques within the direct mail space, which could significantly enhance targeting and personalization strategies. Methodologies deployed through direct mail are the focus of this paper.

In this paper, we propose a supervised learning approach for identifying new customers, i.e., prospecting, which comprises how we define labels for our data and rank potential customers. The casting of prospecting to a supervised learning problem leads to imbalanced tabular data. The current state-of-the-art approach for tabular data is an ensemble of tree-based methods like random forest and XGBoost. We propose a deep learning framework for tabular imbalanced data. This framework is designed to tackle large imbalanced datasets with vast number of numerical and categorical features. Our framework comprises two components: an autoencoder and a feed-forward neural network. We demonstrate the effectiveness of our framework through a transparent real-world case study of prospecting in direct mail advertising. Our results show that our proposed deep learning framework outperforms the state of the art tree-based random forest approach when applied in the real-world.

\end{abstract}


\begin{CCSXML}
<ccs2012>
   <concept>
       <concept_id>10010147.10010257.10010293.10010294</concept_id>
       <concept_desc>Computing methodologies~Neural networks</concept_desc>
       <concept_significance>500</concept_significance>
       </concept>
 </ccs2012>
\end{CCSXML}

\ccsdesc[500]{Computing methodologies~Neural networks}

\keywords{neural networks, imbalanced tabular datasets, direct mail prospecting}



\maketitle

\section{Introduction}
\label{sec:introduction}

Companies and brands use advertising for different purposes; acquiring new customers, creating brand awareness, introducing new products , and so forth. To achieve these goals, advertisers use different mediums such as email, TV, newspaper, direct mail, online advertisement, etc. Each of these techniques has its own pros and cons.  Email advertising campaigns and surveys are less costly to send but have lower success compared to say for example direct mail advertising and surveys~\cite{shih2008comparing,dooley2022paper}.
%\textbf{maybe need a good citation}. 
%BILLY - Potential sources to back this claim include https://journals.sagepub.com/doi/abs/10.1177/1525822X08317085, https://www.baesman.com/news-insights/wait-a-minute-mr.-postman-comparing-direct-mail-marketing-to-email-and-sms-in-2023, https://www.datarobot.com/use-cases/maximize-conversion-rates-for-direct-mail/#:~:text=On%20average%2C%20the%20Direct%20Marketing,a%20high%20customer%20acquisition%20cost, MAYBE https://link.springer.com/article/10.1007/s10900-011-9455-6
%
Previous research has shown that direct mail advertisement is better for brand recall since it requires less cognitive effort to process than digital media~\cite{dooley2022paper}.  We focus on direct mail advertising in this paper.

Direct mail has two types of advertising: \textbf{customer relationship management (CRM)} and \textbf{prospecting}. Customer relationship management uses ads to stay in touch with and motivate customers to make future purchases. The goal when prospecting is to find new customers for the company by using targeted ads. Potential customers are unknown when prospecting, therefore the potential universe that ads can be served to is large (e.g., in the US, we have around $131.43$ million households~\cite{household2022}). Budget limitations and the desire for profitable ad campaigns will necessitate a more targeted approach. Therefore, it is essential to have an efficient framework to acquire new customers with the minimum budget spent. 

Even though direct mail advertising is one of the most effective ways to acquire new customers, to the best of our knowledge, machine learning has not been widely used and studied in direct mail advertising. In this paper, we focus on \textit{direct mail advertising}, specifically on \textbf{prospecting}. Prospecting is challenging since an advertiser must identify new potential customers and target them. To identify and target new customers, we propose an approach that models the problem as a supervised learning task. In our proposed framework, we discuss in detail how we define labels and rank the customers throughout the population. Our proposed approach has delivered strong production performance for a variety of different Fortune 500 companies. Casting the problem of direct mail advertising as a supervised learning problem results in a tabular imbalanced input dataset (which is when the size of different classes or groups are not equal). The majority of classes' distributions can dominate the training process, which will result in poor performance for minority classes.

Previous research corroborated the fact that random forest, XGBoost, and ensemble methods are the first choice in practice and industry due to their simplicity and better performance when working with tabular data~\cite{grinsztajn2022tree}. In addition to using one of these known methods for inference on tabular data (random forest), we propose our deep learning architecture for inference on tabular data. 

%An effective machine learning algorithm for direct mail ads typically aims for efficient customer acquisition vs dollars spent.

%There is widespread use of machine learning in all spheres of our lives including advertising, health, self-driving cars, and entertainment. Deep learning has led to tremendous progress in these spheres. While deep learning has dominated image, audio, and language modeling it has struggled to garner the same usage with tabular data. Among practitioners and data scientists random forest, XGBoost, and ensemble methods are the first choice in practice and industry due to their simplicity and better performance for typical tabular data~\cite{grinsztajn2022tree}.

Developing state-of-the-art deep learning methods for tabular data is an active area of research. While performance benchmarks are common for image, language, and audio data, they are not as readily available for tabular data. There have been some ad-hoc attempts to propose new approaches, such as~\cite{somepalli2021saint,huang2020tabtransformer}. These methods lack extensive evaluation for different datasets and domains. Tabular data also does not have large real-world datasets, as in image and language, to test models. Publicly available datasets are small and do not reflect real-world scenarios.

%Data related to real-world scenarios often contains biases. The most notable being an imbalanced dataset which is when the size of different classes or groups are not equal. The majority classes' distributions can dominate the training process which will result in poor performance for minority classes. Any deep learning method for tabular data should consider these biases. 

We propose a deep learning framework for tabular imbalanced data. We show that its performance is state-of-the-art for tabular data, matching the performance of random forest for the case of direct mail advertising. We also investigate how our proposed solution works in real-world applications of ad serving in the context of direct mail.

Our proposed framework is general and can be applied to any binary classification task for tabular imbalanced data and can be used for both numerical and categorical data. We will illustrate how we cast the problem of prospecting new customers as a binary classification and provide a natural way we can define labels. Finally, we investigate a real-world case study of prospecting via direct mail. In our real-world case study, we show how our proposed framework outperforms an optimized tree-based model over several campaigns.

The rest of this paper is organized as follows:
In Section~\ref{sec:related}, we go over the literature on deep learning models for tabular data. In Section~\ref{sec:Framework}, we propose how we model this problem as a supervised learning task and our deep learning framework, which is followed by our experimental results and real-world implementation in Section~\ref{sec:experiment}. In Section~\ref{sec:discussion}, we discuss the proposed model as well as its limitations and underlying assumptions. Finally, we conclude the paper in Section~\ref{sec:conclusion}.

\section{Related Work}
\label{sec:related}

Deep learning methods for tabular data can be divided into three categories: data transformation, specialized architectures, and regularization techniques~\cite{borisov2022deep}. Data transformation methods convert categorical and numerical data to an input that a deep learning model can effectively process. This can include transforming tabular data into a format similar to an image that captures spatial dependencies, as demonstrated by the successful use of convolutional neural networks in image analysis~\cite{zhu2021converting,sun2019supertml}. This approach has been applied to gene expression profiles and molecular descriptions of drugs, both of which have feature dependencies. However, it may not work well for independent features or when dependencies cannot be captured. Another approach is to use an encoder representation of data such as VIME~\cite{yoon2020vime} to find a new and informative representation of the data for predictive tasks. This has been shown to improve performance compared to baseline models like XGBoost~\cite{chen2016xgboost}, but is limited to continuous data~\cite{somepalli2021saint}. In our framework, we use both the encoder representation and raw representation of data. Note that before now all of these methods have only been tested on small datasets and not yet shown to be scalable to real-world applications.

The majority of the work in this field is concentrated in specialized architectures for tabular data. One approach combines classic machine learning methods, often decision trees, with neural networks~\cite{cheng2016wide,frosst2017distilling,luo2020network,guo2017deepfm}. For example, Frosst and Hinton~\cite{frosst2017distilling} use the outputs of a deep neural network and the ground truth labels to train soft decision trees, which are highly interpretable but sacrifice accuracy. Another specialized architecture is network-on-network (NON)~\cite{luo2020network} which consists of three components: a field-wise network, an across-field network, and an operation fusion network. Each column has its own field-wise network to extract specific information. The optimal operations are chosen by the across-field network and connected by the operation fusion network. For datasets with many features, training individual networks for every column and optimizing operations can be computationally intensive and time-consuming.

Transformers, previously used successfully in other domains, have been applied to tabular data as well. One such architecture is TabNet~\cite{arik2021tabnet}. TabNet uses sequential decision steps to encode features and determine relevant ones for each sample using sparse learned masks. Despite good performance, some inconsistencies in explanations have been noted~\cite{borisov2022deep}. 

The third category of methods for tabular data focuses on regularization techniques~\cite{valdes2021lockout,fiedler2021simple,lounici2021muddling,shavitt2018regularization}. Regularization is necessary due to the extreme flexibility of deep learning models for tabular data that can sometimes lead to overfitting. A Regularization Learning Network (RLN)~\cite{shavitt2018regularization} was proposed to address the observation that a single feature in tabular data can significantly impact the prediction. A RLN is made more efficient and sparse by using regularization coefficients to control the weight of each feature. This approach uses a new "counterfactual loss" to improve performance and the resulting sparse network can also be used for feature importance analysis. The evaluation of a RLN relies mainly on numerical datasets and does not address categorical data. 

In many real-world datasets such as the prospecting case in our study, the number of features and samples are extensive and contain both categorical and numerical data. Additionally, imbalanced datasets are often encountered in these problems. Hence, using specialized regularization techniques for tabular data is not appropriate. Furthermore, due to the large size of these datasets, we aim to implement a fast-training architecture that maintains high performance. 
% Billy - Suggesting a change from
% "While one-time training might be feasible for some tasks, in prospecting, for each company, a new model must be trained from scratch independently."
% to
% "While one-time training might be feasible for some tasks, training a prospecting model must occur independently and from scratch for each company." Sadegh: Done
While one-time training might be feasible for some tasks, training a prospecting model must occur independently and from scratch for each company. 
To address these challenges we present our framework utilizing an autoencoder and feed-forward prediction network.

\section{Proposed Framework}
\label{sec:Framework}

In this section, we present our machine learning approach for prospecting in advertising by formulating it as a supervised learning problem and introducing our deep learning architecture.

Our proposed framework is based on deep learning for tabular data, which has been compared to tree-based methods such as random forests in previous studies, e.g.,~\cite{grinsztajn2022tree}. Deep learning approaches have yet to outperform tree-based methods for small to medium-sized datasets but our problem of prospecting has some key differences to previous theoretical work. First, our datasets are much larger. Second, our datasets are imbalanced, whereas previous conclusions were made from balanced datasets. Third, a deep learning framework provides more options for future optimizations in prospecting, such as fine-tuning the model with transfer learning. 

\subsection{Prospecting}
\label{sub:prospecting}

The goal of prospecting is twofold: first, to find new customers for a company, and second, to prioritize potential customers in an order that minimizes the cost of the prospecting campaign. To achieve these goals, we propose a supervised learning solution.
 
Each company has a record of its previous customers who have purchased one of their products. For each of these customers, a set of features can be created, ranging from demographic information to spending habits and interests. The set of features is both numerical and categorical. This list of customers and their features is referred to as an \textbf{audience}. The type of features that are useful for one product or service being advertised may differ from another. Although creating these features for a customer is not the focus of this paper, some companies, such as Epsilon~\cite{epsilon2023}, Visa~\cite{visa2022}, Mastercard~\cite{mastercard2023}, Acxiom~\cite{acxiom2022}, and Experian~\cite{experian2022} provide such data while abiding by privacy laws and best practices. Further discussion on privacy considerations can be found in Section~\ref{sub:privacy}.

The process of creating features for a company's universe, or the population it is targeting for ads, is similar to creating features for its previous customers. The universe can vary based on the company's marketing goals, ranging from a city to a country or multiple countries. In this framework, the entire US population is used as an example in experimental results. However, this framework is general and not limited by the size of the universe. The population, after excluding the company's existing audience of customers, consists of a mix of potential new customers and individuals who are not as likely to become new customers. The size of the population is usually much larger compared to the audience, with the US population being around three hundred million while the size of the audience is typically not more than one million consumers. 

We model targeted advertisements as a binary classification problem. Therefore, we define two classes: Class 1, which consists of customers in the company's audience list, and Class 0, which consists of a sample of the population minus the audience. We use these two classes to train a model, which is then used to classify the rest of the population and determine the target audience for the ads. It is possible that some individuals in the Class 0 sample are actually potential customers and should be labeled as Class 1. However, due to the small proportion of customers in the population, the number of incorrect Class 0 labels is negligible and is not expected to significantly impact the performance of our proposed framework. The important aspect of our Class 0 labels is that they represent a sample of the universe. 

When we sample data from our population minus our audience, there are two considerations:
\begin{itemize}
    \item Sample size
    \item Sampling method.
\end{itemize}

We aim to balance the size of our sample with the population. On one hand, the sample needs to be large enough to be representative of the non-customer portion of the population. On the other hand, a sample that is too large may contain an increased number of instances with the wrong negative label. This would result in a highly imbalanced dataset that is harder to train a supervised learning model on. We choose a sample size that strikes a balance between being representative and not so large the quantity of false negatives approaches the size of the audience. The actual sample size is chosen through experimentation while taking into account the imbalance in the dataset.

% Billy - should we call this "population minus the audience" its own term? Maybe the "negative set" or "0 class" as it is defined above. -Sadegh: I still prefer the population minus audience, since negative set or 0 class are what we called when sample from populatiom minus audience
There are several methods to sample from the population minus the audience. In this paper we choose to sample uniformly from the entire population without any restrictions.

\subsection{Framework}
\label{sub:framework}

\begin{figure}[h]
    \centering
    \includegraphics[width=0.35\textwidth]{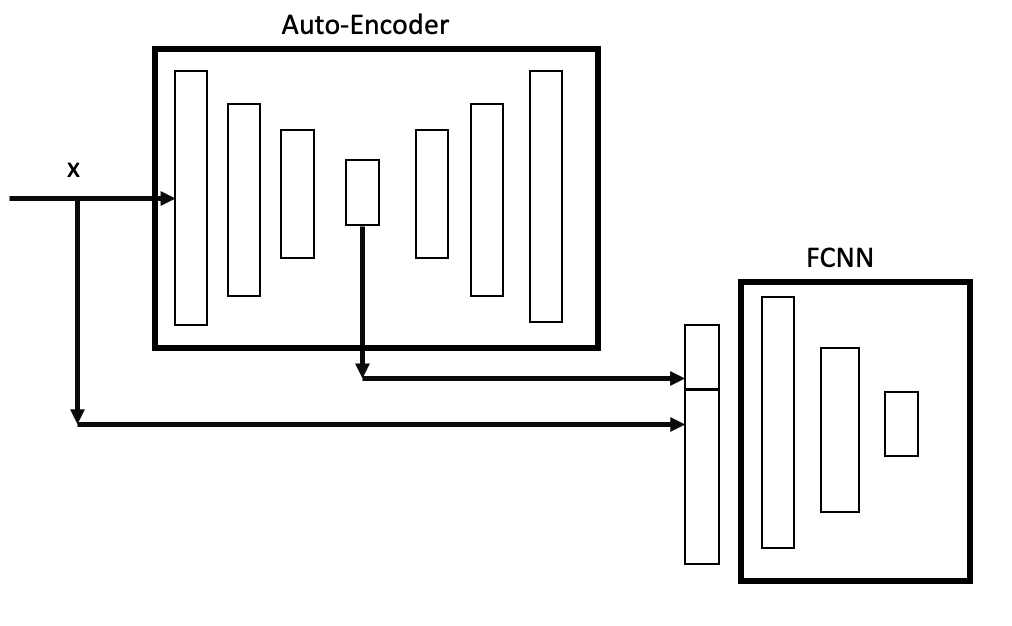}
    \caption{Deep learning framework for tabular data}
    \label{fig:framework}
\end{figure}

In this subsection, we introduce our proposed deep-learning architecture for tabular data. Figure~\ref{fig:framework} shows the detail of our proposed architecture. Our proposed architecture consists of two main components: (1) the autoencoder creates an encoded representation of data to distinguish between the minority class and the majority class, and (2) the feed-forward prediction neural network that uses the combination of encoded representation of data and data itself for final prediction. It is worth mentioning that in our feed-forward prediction neural network, we use a customized loss function for our imbalanced dataset, which we cover in more detail in Section~\ref{subsub:ffn}. 

\subsubsection{Autoencoder}
\label{subsub:auto}
An autoencoder is a type of unsupervised neural network designed to learn an efficient representation of unlabeled data. It consists of two parts: an encoding network and a decoding network.

Encoding: The encoding network takes in the input sample $x$ and compresses it into a lower dimensional representation $x_e$ through a series of linear and non-linear transformations.

\begin{equation}
    x_e = f_e(x; \theta_e), 
\end{equation}

where $f_e$ is encoding neural network with weight parameters $\theta_e$.

Decoding: The decoding network then takes the encoded representation $x_e$ and tries to reconstruct the original input $x$ by applying a reverse process of linear and non-linear transformations.

\begin{equation}
    x' = f_d(x_e; \theta_d),   
\end{equation}

where $x'$ is the reconstructed output and $f_d$ is decoding neural network with weight parameters $\theta_d$.

The objective of the autoencoder is to minimize the reconstruction error between the original input and the reconstructed output. The autoencoder achieves that by minimizing the distance between $x$ and $x'$.

\begin{equation}
    min_{\theta_e, \theta_d} dist(x, x'),
\end{equation}

where $dist(*)$ in this case is a measure of  Euclidean distance.

This process helps in learning compact and meaningful representations of the input data that can be useful for further analysis. In our case, we use the learned encoding to train a supervised learning model for the task of targeted direct mail advertisements.

\subsubsection{Feed-Forward Prediction Network}
\label{subsub:ffn}

Our feed-forward neural network uses the concatenation of our raw features and the encoded version of these features as input. This concatenated input is then used to build a classification model to predict the final class label.

\begin{equation}
    z = [x, x_e]
\end{equation}
\begin{equation}
    \hat{y} = f_{fnn}(z; \theta_m)
\end{equation}

where $z$ denotes the input features to the feed-forward neural network, i.e., the concatenation of input features and the encoded features. The weight parameters of the feed-forward neural network are represented by $\theta_m$, and the predicted probability of label $1$ is represented by $\hat{y}$. 

\textbf{Cost Function.} Imbalanced data, where different classes have varying sizes, is a common challenge in machine learning. There are several approaches to address this issue, such as re-sampling or cost-sensitive mechanisms. Over-sampling involves repeating samples from the minority class or generating new samples via methods like interpolating neighboring samples~\cite{chawla2002smote} or data augmentation using generative models~\cite{antoniou2017data}. Other approaches involve assigning weights to each sample's loss based on the data distribution~\cite{kahn1953methods,khan2017cost}, and the most common method is to select weights for each class as the inverse of the class frequency~\cite{huang2016learning}. A more advanced approach to weight selection involves learning weights from a balanced sample of data~\cite{kamani2020targeted}. In this paper, we focus on the latter approach of assigning weights to the sample's loss function and using the inverse of the class frequency for weight selection.

Since we are modeling the prospecting scenario as a classification problem, we use binary cross entropy for our loss function. During training, our objective is to minimize the following loss function:

\begin{equation}
    - \frac{1}{N} \sum_{i=1}^{N} w_1 y_i \log(p_i) + w_0 (1-y_i) \log(1 - p_i)
\end{equation}

where $N$ is the number of samples in training. $y_i$ is the actual label of sample $i$ and $p_i$ is the model prediction probability for sample $i$. Note that $w_0$ and $w_1$ are the weight parameters of class $0$ and class $1$, respectively, which we choose as the inverse of the class frequency. 

%and finding the weight parameter experimentally. However, in our experiment, we narrow our search for the best weight parameter considering that choosing weight as the inverse of class frequency usually yields good results. 

\subsection{Random Forest vs Our Framework}

The current state-of-the-art approach for classification in tabular data favors tree-based models such as random forest and XGBoost for small to medium datasets (around 10K samples)~\cite{grinsztajn2022tree}. However, in many practical applications, the datasets are much larger, often exceeding 100K samples and sometimes reaching millions, as is the case in our study of prospecting via direct mail. In this study, we compare the real-world performance of our proposed deep learning framework, consisting of an autoencoder and feed-forward neural network, to a random forest model regarding conversion rate and found that our framework outperformed the random forest.

As mentioned earlier, tree-based algorithms provide state-of-the-art performance for tabular data~\cite{grinsztajn2022tree}. In~\cite{grinsztajn2022tree}, Grinsztajn et al. compare tree-based algorithms like random forest with other approaches and different datasets thoroughly and show that tree-based approaches outperform deep learning and other traditional methods.  
As a result of this previous work, we only compare our proposed framework to the random forest model in our evaluation.

\section{Experimental Results}
\label{sec:experiment}

In this section, we present the experimental results of our proposed framework for addressing the issue of prospecting in imbalanced datasets using real-world data. First, we provide a brief overview of our datasets, including their creation and evaluation criteria. Next, we describe the process of finalizing the proposed model structure and hyperparameter tuning through experimentation. We also discuss the selection of sample size. Finally, we evaluate the performance of our proposed model in a real-world scenario.

\subsection{Dataset}
\label{sub:dataset}

Each country has its own privacy laws. Since we are a company based in the United States, we restrict ourselves to the United States' privacy laws. However, our proposed method is general and can be used in other countries.  The data we use for our experiments is provided by companies like Acxiom and Experian, which offer features for each individual or household and are compliant with federal and state privacy laws and best practices. To protect the privacy of individuals, the data provided was de-identified during the model development process. For a further discussion on privacy, see Section~\ref{sub:privacy}.

While these data providers supply the features for our input, they do not provide the labels for our binary classification. As mentioned earlier, each company seeking to do prospecting provides a list of their customers, i.e., the audience. We use this list to create the label 1 data by combining it with the features provided by data providers like Acxiom~\cite{acxiom2022} and Experian~\cite{experian2022}. To create the label 0 data, we sample from the population that is not included in the audience list. By combining both label 0 and label 1 data, we create our dataset. The term "ratio" is used to denote the sample size in relation to the audience size. For example, a ratio of 4 means that the sample size is equal to 4 times the audience size.

In our experimental results, we use audience lists from six companies, referred to as $A$, $B$, $C$, $D$, $E$, and $F$, to build our dataset, train the model, and evaluate the proposed model. However, for confidentiality reasons, the actual names of these six advertisers are not disclosed. In the following sections, we provide a general description of these companies without revealing their specific identities.

\begin{itemize}
    \item Company $A$: An apparel company with an audience size of $107,098$. 
    
    \item Company $B$: An apparel company mainly for women specialized for swim and fitness with an audience size of $47,334$.
    
    \item Company $C$: A health related company and with an audience size of $22,994$.
    
    \item Company $D$: A sustainable lifestyle essentials brand for women with an audience size of $20,035$.
    
    \item Company $E$: A home services company with an audience size of $390,273$.
    
    \item Company $F$: A home decor company with an audience size of $446,606$.
\end{itemize}

Note that the audiences of these six companies vary in terms of both size and product type. We select these companies to demonstrate the effectiveness of our proposed framework across different product verticals. To show how this framework can be generalized, we use companies $A$, $B$, $C$, and $D$ for training, model building, and hyperparameter tuning. Then, we evaluate the performance of our finalized framework in a real-world scenario using companies $E$ and $F$.

For consistency, we use the same data provider throughout the experiment. However, we do not disclose which provider we used among those mentioned. 

We evaluate our proposed framework in two ways. First, we use traditional evaluation metrics such as accuracy, precision, recall, and the $F_2$ score. Since our data is imbalanced, we focus on precision, recall, and the $F_2$ score, which is a weighted harmonic mean of precision and recall. We split our dataset into training and testing sets, and evaluate the performance of our model based on these metrics. Recall is particularly important when prospecting because we do not want to miss any potential customers. Therefore, we consider the $F_2$ score which places more emphasis on recall.

Second, we evaluate our proposed framework in a real-world direct mail scenario. Based on our trained model, we measure the number of consumers exposed to a direct mail ad who then make a purchase within a specific time frame (attribution window). These results are presented in Section~\ref{sub:real-world}. 

Our proposed deep learning structure consists of two parts: an autoencoder and a feed-forward neural network. To train the framework, we first train the autoencoder alone to ensure that the encoder produces good compact representations of the input features. Then, we use the trained autoencoder in the framework and freeze its weights during the training of the feed-forward neural network.

This separate training and freezing of the autoencoder has several benefits. First, it is faster to train the autoencoder alone than it is to simultaneously train both the feed-forward neural network and the autoencoder using a joint loss function. Second, we found that using the pre-trained autoencoder and fine-tuning it during the training process of the feed-forward neural network does not result in improved performance. Finally, jointly training the autoencoder and the feed-forward neural network would introduce an additional hyper-parameter to tune, i.e., a weight parameter specifying the contribution of each loss function, which we aim to avoid for simplicity.

\subsection{Architecture and Hyper-parameters}
\label{sub:architecture}

\begin{table*}[h]
\footnotesize
\begin{tabular}{|cccccc|cccccc|}
\hline
\multicolumn{6}{|c|}{Company A}                                                                                                                                                & \multicolumn{6}{c|}{Company B}                                                                                                                                                \\ \hline
\multicolumn{1}{|c|}{Size} & \multicolumn{1}{c|}{Train-Test} & \multicolumn{1}{c|}{Accuracy} & \multicolumn{1}{c|}{Precision} & \multicolumn{1}{c|}{Recall}  & $F_2$   & \multicolumn{1}{c|}{Size} & \multicolumn{1}{c|}{Train-Test} & \multicolumn{1}{c|}{Accuracy} & \multicolumn{1}{c|}{Precision} & \multicolumn{1}{c|}{Recall}  & $F_2$   \\ \hline
\multicolumn{1}{|c|}{16}           & \multicolumn{1}{c|}{Train}      & \multicolumn{1}{c|}{$71.69 \pm 0.09$}  & \multicolumn{1}{c|}{$39.6 \pm 0.09$}   & \multicolumn{1}{c|}{$79.05 \pm 0.1$} & $65.91 0.07$ & \multicolumn{1}{c|}{16}           & \multicolumn{1}{c|}{Train}      & \multicolumn{1}{c|}{$84.74 \pm 0.12$}  & \multicolumn{1}{c|}{$57.79 \pm 0.24$}   & \multicolumn{1}{c|}{$88.05 \pm 0.26$} & $79.7 \pm 0.1$ \\ \hline
\multicolumn{1}{|c|}{16}           & \multicolumn{1}{c|}{Test}       & \multicolumn{1}{c|}{$70.88 \pm 0.27$}  & \multicolumn{1}{c|}{$38.51 \pm 0.28$}   & \multicolumn{1}{c|}{$76.6 \pm 0.34$} & $63.95 \pm 0.12$ & \multicolumn{1}{c|}{16}           & \multicolumn{1}{c|}{Test}       & \multicolumn{1}{c|}{$84.14 \pm 0.62$}  & \multicolumn{1}{c|}{$57.21 \pm 1.17$}   & \multicolumn{1}{c|}{$85.11 \pm 1.06$} & $77.53 \pm 0.33$ \\ \hline
\multicolumn{1}{|c|}{32}           & \multicolumn{1}{c|}{Train}      & \multicolumn{1}{c|}{$71.89 \pm 0.43$}  & \multicolumn{1}{c|}{$39.77 \pm 0.35$}   & \multicolumn{1}{c|}{$78.74 \pm 0.62$} & $65.83 \pm 0.16$ & \multicolumn{1}{c|}{32}           & \multicolumn{1}{c|}{Train}      & \multicolumn{1}{c|}{$84.55 \pm 0.19$}  & \multicolumn{1}{c|}{$57.34 \pm 0.37$}   & \multicolumn{1}{c|}{$88.59 \pm 0.33$}  & $79.88 \pm 0.11$ \\ \hline
\multicolumn{1}{|c|}{32}           & \multicolumn{1}{c|}{Test}       & \multicolumn{1}{c|}{$70.89 \pm 0.34$}  & \multicolumn{1}{c|}{$38.52 \pm 0.4$}   & \multicolumn{1}{c|}{$76.6 \pm 0.65$} & $63.96 \pm 0.24$ & \multicolumn{1}{c|}{32}           & \multicolumn{1}{c|}{Test}       & \multicolumn{1}{c|}{$83.27 \pm 0.4$}  & \multicolumn{1}{c|}{$55.47 \pm 0.77$}   & \multicolumn{1}{c|}{$86.44 \pm 0.59$}  & $77.75 \pm 0.26$ \\ \hline
\multicolumn{1}{|c|}{64}           & \multicolumn{1}{c|}{Train}      & \multicolumn{1}{c|}{$72.03 \pm 0.19$}  & \multicolumn{1}{c|}{$39.91 \pm 0.17$}   & \multicolumn{1}{c|}{$78.76 \pm 0.25$} & $65.93 \pm 0.08$ & \multicolumn{1}{c|}{64}           & \multicolumn{1}{c|}{Train}      & \multicolumn{1}{c|}{$84.62 \pm 0.13$}  & \multicolumn{1}{c|}{$54.47 \pm 0.28$}   & \multicolumn{1}{c|}{$88.59 \pm 0.3$} & $79.93 \pm 0.12$ \\ \hline
\multicolumn{1}{|c|}{64}           & \multicolumn{1}{c|}{Test}       & \multicolumn{1}{c|}{$70.92 \pm 0.28$}  & \multicolumn{1}{c|}{$38.51 \pm 0.18$}   & \multicolumn{1}{c|}{$76.59 \pm 0.58$} & $63.94 \pm 0.28$  & \multicolumn{1}{c|}{64}           & \multicolumn{1}{c|}{Test}       & \multicolumn{1}{c|}{$83.86 \pm 0.43$}  & \multicolumn{1}{c|}{$56.61 \pm 0.77$}   & \multicolumn{1}{c|}{$85.65 \pm 0.54$} & $77.67 \pm 0.19$ \\ \hline
\multicolumn{1}{|c|}{128}          & \multicolumn{1}{c|}{Train}      & \multicolumn{1}{c|}{$71.99 \pm 0.27$}  & \multicolumn{1}{c|}{$39.9 \pm 0.25$}   & \multicolumn{1}{c|}{$79.07 \pm 0.34$} & $66.09 \pm 0.07$ & \multicolumn{1}{c|}{128}          & \multicolumn{1}{c|}{Train}      & \multicolumn{1}{c|}{$84.9 \pm 0.16$}   & \multicolumn{1}{c|}{$58.01 \pm 0.33$}   & \multicolumn{1}{c|}{$88.43 \pm 0.24$} & $80.04 \pm 0.09$ \\ \hline
\multicolumn{1}{|c|}{128}          & \multicolumn{1}{c|}{Test}       & \multicolumn{1}{c|}{$71.03 \pm 0.26$}  & \multicolumn{1}{c|}{$38.64 \pm 0.14$}   & \multicolumn{1}{c|}{$76.46 \pm 0.25$} & $63.95 \pm 0.21$  & \multicolumn{1}{c|}{128}          & \multicolumn{1}{c|}{Test}       & \multicolumn{1}{c|}{$83.59 \pm 0.76$}  & \multicolumn{1}{c|}{$56.12 \pm 0.14$}   & \multicolumn{1}{c|}{$85.98 \pm 1.07$} & $77.69 \pm 0.22$ \\ \hline
\multicolumn{6}{|c|}{Company C}                                                                                                                                                & \multicolumn{6}{c|}{Company D}                                                                                                                                                \\ \hline
\multicolumn{1}{|c|}{Size} & \multicolumn{1}{c|}{Train-Test} & \multicolumn{1}{c|}{Accuracy} & \multicolumn{1}{c|}{Precision} & \multicolumn{1}{c|}{Recall}  & $F_2$   & \multicolumn{1}{c|}{Size} & \multicolumn{1}{c|}{Train-Test} & \multicolumn{1}{c|}{Accuracy} & \multicolumn{1}{c|}{Precision} & \multicolumn{1}{c|}{Recall}  & $F_2$   \\ \hline
\multicolumn{1}{|c|}{16}           & \multicolumn{1}{c|}{Train}      & \multicolumn{1}{c|}{$72.39 \pm 0.31$}  & \multicolumn{1}{c|}{$40.21 \pm 0.26$}   & \multicolumn{1}{c|}{$78.08 \pm 0.63$} & $65.70 \pm 0.23$ & \multicolumn{1}{c|}{16}           & \multicolumn{1}{c|}{Train}      & \multicolumn{1}{c|}{$89.67 \pm 0.1$}  & \multicolumn{1}{c|}{$67.62 \pm 0.33$}   & \multicolumn{1}{c|}{$92.85 \pm 0.23$} & $86.4 \pm 0.09$ \\ \hline
\multicolumn{1}{|c|}{16}           & \multicolumn{1}{c|}{Test}       & \multicolumn{1}{c|}{$70.35 \pm 0.45$}  & \multicolumn{1}{c|}{$37.80 \pm 0.6$}   & \multicolumn{1}{c|}{$74.79 \pm 0.75$} & $62.55 \pm 0.37$ & \multicolumn{1}{c|}{16}           & \multicolumn{1}{c|}{Test}       & \multicolumn{1}{c|}{$88.45 \pm 0.21$}  & \multicolumn{1}{c|}{$65.11 \pm 0.71$}   & \multicolumn{1}{c|}{$89.94 \pm 0.48$}  & $83.57 \pm 0.51$ \\ \hline
\multicolumn{1}{|c|}{32}           & \multicolumn{1}{c|}{Train}      & \multicolumn{1}{c|}{$72.35 \pm 0.47$}  & \multicolumn{1}{c|}{$40.19 \pm 0.44$}     & \multicolumn{1}{c|}{$78.35 \pm 0.63$} & $65.85 \pm 0.2$ & \multicolumn{1}{c|}{32}           & \multicolumn{1}{c|}{Train}      & \multicolumn{1}{c|}{$89.68 \pm 0.09$}   & \multicolumn{1}{c|}{$67.64 \pm 0.28$}   & \multicolumn{1}{c|}{$92.87 \pm 0.21$} & $86.42 \pm 0.08$ \\ \hline
\multicolumn{1}{|c|}{32}           & \multicolumn{1}{c|}{Test}       & \multicolumn{1}{c|}{$70.29 \pm 0.81$}  & \multicolumn{1}{c|}{$37.75 \pm 0.72$}   & \multicolumn{1}{c|}{$74.68 \pm 1.34$} & $62.45 \pm 0.51$ & \multicolumn{1}{c|}{32}           & \multicolumn{1}{c|}{Test}       & \multicolumn{1}{c|}{$88.64 \pm 0.16$}  & \multicolumn{1}{c|}{$65.58 \pm 0.66$}   & \multicolumn{1}{c|}{$89.88 \pm 0.6$} & $83.68 \pm 0.46$  \\ \hline
\multicolumn{1}{|c|}{64}           & \multicolumn{1}{c|}{Train}      & \multicolumn{1}{c|}{$72.13 \pm 0.17$}  & \multicolumn{1}{c|}{$40.01 \pm 0.19$}   & \multicolumn{1}{c|}{$78.83 \pm 0.2$} & $66.02 \pm 0.08$  & \multicolumn{1}{c|}{64}           & \multicolumn{1}{c|}{Train}      & \multicolumn{1}{c|}{$89.7 \pm 0.06$}  & \multicolumn{1}{c|}{$67.66 \pm 0.22$}   & \multicolumn{1}{c|}{$92.94 \pm 0.12$} & $86.48 \pm 0.07$ \\ \hline
\multicolumn{1}{|c|}{64}           & \multicolumn{1}{c|}{Test}       & \multicolumn{1}{c|}{$70.56 \pm 0.1$}  & \multicolumn{1}{c|}{$37.98 \pm 0.35$}   & \multicolumn{1}{c|}{$74.61 \pm 0.58$} & $62.54 \pm 0.47$ & \multicolumn{1}{c|}{64}           & \multicolumn{1}{c|}{Test}       & \multicolumn{1}{c|}{$88.71 \pm 0.26$}  & \multicolumn{1}{c|}{$65.81 \pm 0.71$}   & \multicolumn{1}{c|}{$89.73 \pm 0.72$} & $83.64 \pm 0.4$ \\ \hline
\multicolumn{1}{|c|}{128}          & \multicolumn{1}{c|}{Train}      & \multicolumn{1}{c|}{$72.31 \pm 0.14$}  & \multicolumn{1}{c|}{$40.21 \pm 0.1$}   & \multicolumn{1}{c|}{$78.97 \pm 0.33$}  & $66.21 \pm 0.17$ & \multicolumn{1}{c|}{128}          & \multicolumn{1}{c|}{Train}      & \multicolumn{1}{c|}{$89.73 \pm 0.15$}   & \multicolumn{1}{c|}{$67.71 \pm 0.43$}   & \multicolumn{1}{c|}{$93.15 \pm 0.28$} & $86.63 \pm 0.11$ \\ \hline
\multicolumn{1}{|c|}{128}          & \multicolumn{1}{c|}{Test}       & \multicolumn{1}{c|}{$70.82 \pm 0.66$}  & \multicolumn{1}{c|}{$38.2 \pm 0.74$}   & \multicolumn{1}{c|}{$74.2 \pm 0.94$} & $62.42 \pm 0.32$ & \multicolumn{1}{c|}{128}          & \multicolumn{1}{c|}{Test}       & \multicolumn{1}{c|}{$88.35 \pm 0.45$}  & \multicolumn{1}{c|}{$64.86 \pm 0.98$}   & \multicolumn{1}{c|}{$90.19 \pm 0.7$} & $83.65 \pm 0.46$ \\ \hline
\end{tabular}
\caption{Comparison of different encoder size for these four companies}
\label{tab:company_encoder_size}
\end{table*}

Our architecture comprises two parts: an autoencoder and a feed-forward neural network. First, we determine the optimal encoded size for the autoencoder. We use a symmetrical structure for the encoder and decoder, with the first hidden layer in the encoder having 256 neurons and each subsequent layer size being halved until we reach the desired encoded size. We then reconstruct the input from the encoded output and compare the decoder output to the original input using Euclidean distance. We have 734 input features, so an encoded representation less than 16 may not be sufficient and a representation larger than 128 may not be compact enough. Thus, we consider encoded output sizes of 16, 32, 64, and 128. We compare the precision, recall, and $F_2$ measures for these four sizes and choose the size with the best performance without overfitting. For the training of our autoencoder, we use the Adam optimizer and train for 100 epochs until the distance metric plateaus. Based on the results, we select an encoded size of 32 for our encoder. The results for the four sizes are shown in Table~\ref{tab:company_encoder_size}. We choose an encoded size of 32 as the $F_2$ measure is slightly better for this size.

%For each of the four companies, the highest $F_2$ score and recall value in the test set is for an encoder of size $32$. However, the values are very close to each other and for each of these four companies, the difference in terms of metrics in the training set and the test set is negligible. In other words, the overfitting level for different encoder sizes is nearly the same.

Our proposed deep learning framework shows different levels of performance for different companies in terms of recall and precision. For example, companies $A$ and $C$ have lower precision values (below $50\%$) compared to companies B and D (above $50\%$). The same is true for recall values, with companies B and D having about $10\%$ higher recall compared to companies A and C. One reason for this variation is the difference in products offered by these companies, some of which have more general products that can be purchased by a larger customer base, making it more challenging to build specific prospecting models. As a result, our framework tends to produce more false positives (lower precision) to ensure that no potential customers are missed (high recall). Additionally, all four companies have an imbalanced dataset with varying class frequencies, which affects the weight assigned to each sample in the loss function. This leads to a reasonable rate of true positives while having a tendency to predict more positives and higher weights for class 1 predictions, resulting in more false positives.

The second component, the feed-forward neural network, takes as input the concatenation of the input features (734) and the output of the encoder (32). We test three different network architectures to find the best one for our problem. 

The first architecture (architecture 512) has 512 neurons in the first layer and each subsequent layer reduces the number of neurons by half until reaching 64 neurons. The second architecture (architecture 2048) starts with 2048 neurons in the first layer and reduces the number of neurons by half in each subsequent layer until 64 neurons. The third architecture (architecture 4096) has 4096 neurons in the first layer followed by 64 neurons. All three architectures have a final layer of 1 neuron for binary prediction. Batch normalization, the ReLU function, and dropout~\cite{srivastava2014dropout} are used in all layers of these three architectures. 
%The performance of these architectures was evaluated in the same manner as the autoencoder evaluation to determine which one provides the best performance while also having low overfitting and reasonable training time. 

Due to space limitations, we do not cover in detail the performance of these three structures. However, the performance of these architectures was evaluated in the same manner as the autoencoder evaluation to determine which one provides the best performance while also having low overfitting and reasonable training time. As a result, we select architecture $4096$ in our framework.

In our model, we use a dropout probability of 0.5. For the optimizer, we test Adam, AdamW~\cite{loshchilov2017decoupled}, and SGD with momentum~\cite{sutskever2013importance}. Based on our results, we observe that SGD with momentum gives the best performance with lower overfitting. We select a learning rate of 0.0001 with a momentum of 0.92. We use a training batch size of 256 and train for 100 epochs. 

\subsection{Ratio}
\label{sub:ratio}
\begin{figure}
     \centering
     \begin{subfigure}{0.4\textwidth}
         \centering
         \includegraphics[width=\textwidth]{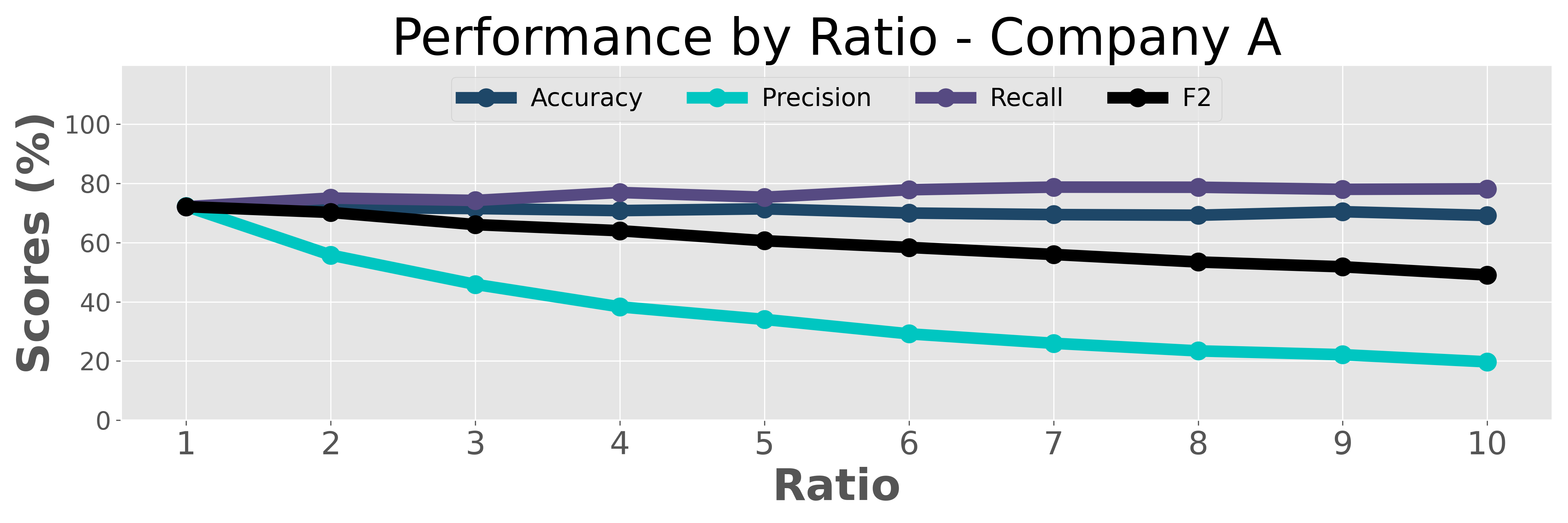}
         \caption{Company $A$}
         \label{fig:comp_a_ratio}
     \end{subfigure}
     \hfill
     \begin{subfigure}{0.4\textwidth}
         \centering
         \includegraphics[width=\textwidth]{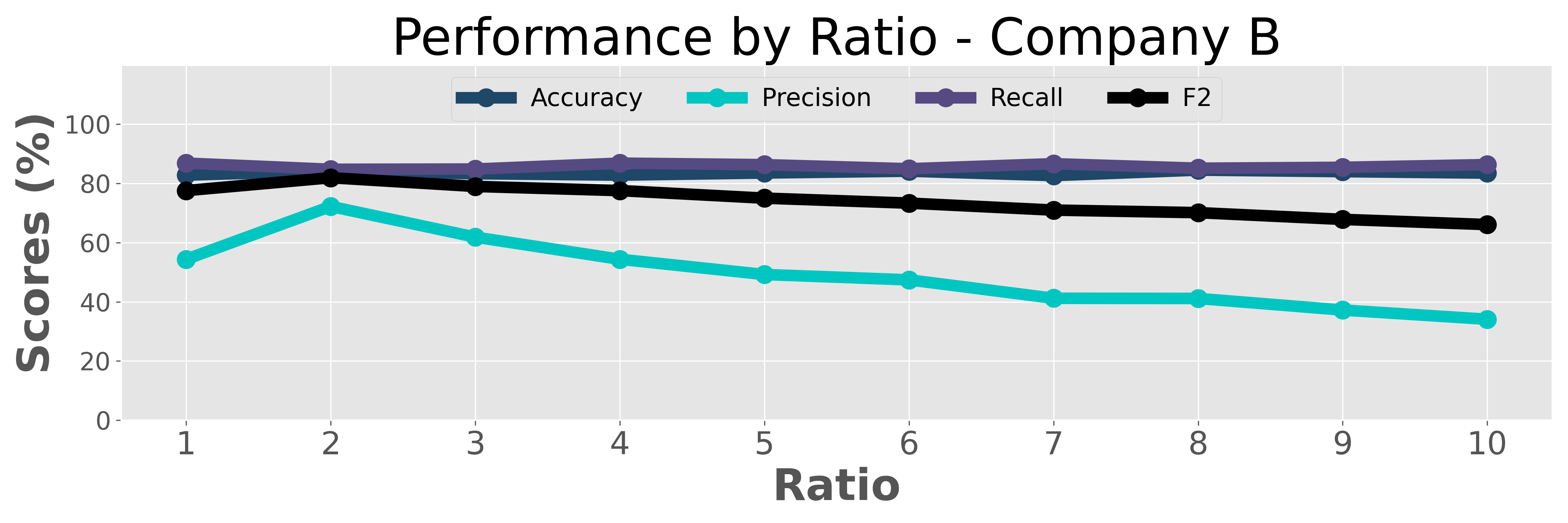}
         \caption{Company $B$}
         \label{fig:comp_b_ratio}
     \end{subfigure}
     \hfill
     \begin{subfigure}{0.4\textwidth}
         \centering
         \includegraphics[width=\textwidth]{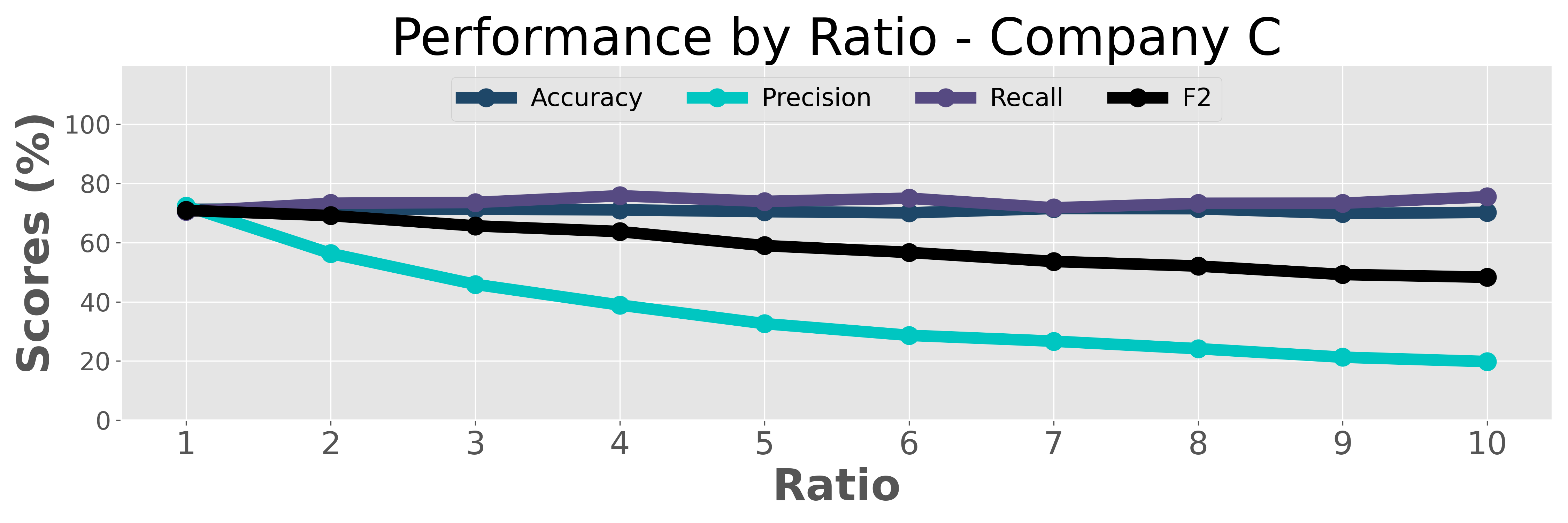}
         \caption{Company $C$}
         \label{comp_c_ratio}
     \end{subfigure}
     \hfill
     \begin{subfigure}{0.4\textwidth}
         \centering
         \includegraphics[width=\textwidth]{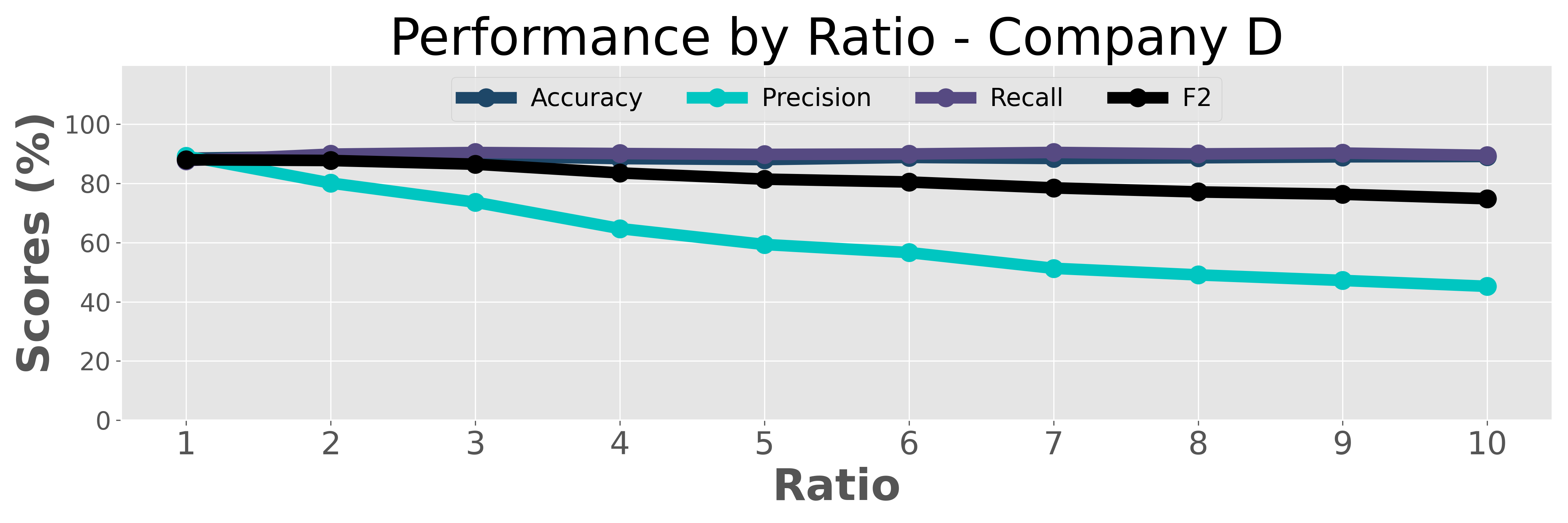}
         \caption{Company $D$}
         \label{comp_d_ratio}
     \end{subfigure}
        \caption{Comparison of different ratios for different companies performance on test set}
        \label{fig:comp_ratios}
\end{figure}

The ratio of the sample size to the target audience size is investigated in this section. The dataset for our prospecting task is created by using a sample of the population excluding the target audience as the negative class (label 0).  We call the size of the sample relative to the audience size the \textit{ratio}. When the ratio is greater than 1, the dataset becomes imbalanced, meaning that the number of samples with label 0 exceeds the number of samples with label 1. To prevent the majority class (label 0) from dominating the learning process and lowering the accuracy of the minority class (label 1), weights are assigned to each sample's loss based on the inverse of the class frequency. The results for this investigation can be seen in Figure~\ref{fig:comp_ratios} (see Tables~\ref{tab:company_A_ratio}, ~\ref{tab:company_B_ratio}, ~\ref{tab:company_C_ratio}, and ~\ref{tab:company_D_ratio} in Appendix~\ref{app:archs} for both training and test results).

According to the results in the figure, as the ratio increases the precision for both the test and training sets decreases. This outcome is expected because as the ratio increases, there are more samples for class with label $0$. The weight factor (inverse of class frequency) in the loss function yields more positive predictions to maintain a high recall value, but also results in more false positives, which is undesirable. However, the more data we use for training the more generalizable we make the model. Therefore, when choosing the ratio, it is important to balance these two factors and opt for a higher ratio as long as the precision does not excessively degrade.

As mentioned previously, ratios $1$ and $2$ are not chosen as they lead to small datasets, which limit the model's ability to generalize. Additionally, precision values for ratios greater than $5$ are too low. Hence, we focus on ratios $3$, $4$, and $5$ and do extensive analysis for these three ratios, see Table~\ref{tab:comarison_candidate} for test performance and Table~\ref{tab:comarison_candidate_train} for training set performance. For companies $A$, $B$, and $C$, the highest recall is achieved with a ratio of $4$. Although company $D$ has better recall with ratio $3$, the difference is minimal and the recall value for ratio $4$ is very close. Despite a slight decrease in precision when increasing the ratio from $3$ to $4$, we still opt for ratio $4$ as it offers a higher recall and a larger dataset, which is more ideal for our real-world testing with variable audience sizes.

\begin{table*}[]
\begin{tabular}{|ccccc|ccccc|}
\hline
\multicolumn{5}{|c|}{Company A}                                                                                                                                        & \multicolumn{5}{c|}{Company B}                                                                                                                                                       \\ \hline
\multicolumn{1}{|c|}{Ratio} & \multicolumn{1}{c|}{Accuracy}         & \multicolumn{1}{c|}{Precision}        & \multicolumn{1}{c|}{Recall}           & $F_2$            & \multicolumn{1}{c|}{Ratio} & \multicolumn{1}{c|}{Accuracy}                        & \multicolumn{1}{c|}{Precision}        & \multicolumn{1}{c|}{Recall}           & $F_2$            \\ \hline
\multicolumn{1}{|c|}{$3$}   & \multicolumn{1}{c|}{$71.11 \pm 0.57$} & \multicolumn{1}{c|}{$45.19 \pm 0.56$} & \multicolumn{1}{c|}{$76.08 \pm 1.37$} & $66.92 \pm 0.6$  & \multicolumn{1}{c|}{3}     & \multicolumn{1}{c|}{$83.48 \pm 0.17$}                & \multicolumn{1}{c|}{$62.26 \pm 0.45$} & \multicolumn{1}{c|}{$85.18 \pm 0.19$} & $79.34 \pm 0.25$ \\ \hline
\multicolumn{1}{|c|}{4}     & \multicolumn{1}{c|}{$71.11 \pm 0.19$} & \multicolumn{1}{c|}{$38.55 \pm 0.55$} & \multicolumn{1}{c|}{$76.57 \pm 0.82$} & $63.94 \pm 0.16$ & \multicolumn{1}{c|}{4}     & \multicolumn{1}{c|}{$83.47 \pm 0.45$}                & \multicolumn{1}{c|}{$55.62 \pm 0.82$} & \multicolumn{1}{c|}{$85.81 \pm 0.73$} & $77.4 \pm 0.18$  \\ \hline
\multicolumn{1}{|c|}{5}     & \multicolumn{1}{c|}{$71.56 \pm 0.35$} & \multicolumn{1}{c|}{$34.27 \pm 0.52$} & \multicolumn{1}{c|}{$75.51 \pm 1.74$} & $60.84 \pm 0.61$ & \multicolumn{1}{c|}{5}     & \multicolumn{1}{c|}{$84.09 0.13$} & \multicolumn{1}{c|}{$52.22 \pm 2.65$} & \multicolumn{1}{c|}{$85.57 \pm 0.59$} & $75.83 \pm 0.92$ \\ \hline
\multicolumn{5}{|c|}{Company C}                                                                                                                                        & \multicolumn{5}{c|}{Company D}                                                                                                                                                       \\ \hline
\multicolumn{1}{|c|}{Ratio} & \multicolumn{1}{c|}{Accuracy}         & \multicolumn{1}{c|}{Precision}        & \multicolumn{1}{c|}{Recall}           & $F_2$            & \multicolumn{1}{c|}{Ratio} & \multicolumn{1}{c|}{Accuracy}                        & \multicolumn{1}{c|}{Precision}        & \multicolumn{1}{c|}{Recall}           & $F_2$            \\ \hline
\multicolumn{1}{|c|}{3}     & \multicolumn{1}{c|}{$71.05 \pm 0.28$} & \multicolumn{1}{c|}{$45.27 \pm 0.39$} & \multicolumn{1}{c|}{$73.15 \pm 0.86$} & $65.12 \pm 0.62$ & \multicolumn{1}{c|}{3}     & \multicolumn{1}{c|}{$88.92 \pm 0.26$}                & \multicolumn{1}{c|}{$72.88 \pm 0.69$} & \multicolumn{1}{c|}{$90.35 \pm 0.13$} & $86.22 \pm 0.28$ \\ \hline
\multicolumn{1}{|c|}{4}     & \multicolumn{1}{c|}{$70.89 \pm 0.28$} & \multicolumn{1}{c|}{$38.71 \pm 0.37$} & \multicolumn{1}{c|}{$74.84 \pm 0.76$} & $63.03 \pm 0.52$ & \multicolumn{1}{c|}{4}     & \multicolumn{1}{c|}{$88.58 \pm 0.16$}                & \multicolumn{1}{c|}{$65.62 \pm 0.84$} & \multicolumn{1}{c|}{$90.24 \pm 0.55$} & $83.94 \pm 0.3$  \\ \hline
\multicolumn{1}{|c|}{5}     & \multicolumn{1}{c|}{$70.03 \pm 0.39$} & \multicolumn{1}{c|}{$32.74 \pm 0.08$} & \multicolumn{1}{c|}{$74.64 \pm 0.77$} & $59.43 \pm 0.4$  & \multicolumn{1}{c|}{5}     & \multicolumn{1}{c|}{$88.5 \pm 0.42$}                 & \multicolumn{1}{c|}{$60.25 \pm 1.41$} & \multicolumn{1}{c|}{$90.27 \pm 0.32$} & $82.08 \pm 0.6$  \\ \hline
\end{tabular}
\caption{In-depth comparison of test performance of candidate ratios of different companies}
\label{tab:comarison_candidate}
\end{table*}

\begin{table*}[]
\begin{tabular}{|ccccc|ccccc|}
\hline
\multicolumn{5}{|c|}{Company A}                                                                                                                                        & \multicolumn{5}{c|}{Company B}                                                                                                                                                       \\ \hline
\multicolumn{1}{|c|}{Ratio} & \multicolumn{1}{c|}{Accuracy}         & \multicolumn{1}{c|}{Precision}        & \multicolumn{1}{c|}{Recall}           & $F_2$            & \multicolumn{1}{c|}{Ratio} & \multicolumn{1}{c|}{Accuracy}                        & \multicolumn{1}{c|}{Precision}        & \multicolumn{1}{c|}{Recall}           & $F_2$            \\ \hline
\multicolumn{1}{|c|}{$3$}   & \multicolumn{1}{c|}{$72.09 \pm 0.1$} & \multicolumn{1}{c|}{$46.55 \pm 0.09$} & \multicolumn{1}{c|}{$78.06 \pm 0.17$} & $68.75 \pm 0.08$  & \multicolumn{1}{c|}{3}     & \multicolumn{1}{c|}{$84.95 \pm 0.13$}                & \multicolumn{1}{c|}{$64.83 \pm 0.32$} & \multicolumn{1}{c|}{$87.14 \pm 0.24$} & $81.52 \pm 0.07$ \\ \hline
\multicolumn{1}{|c|}{4}     & \multicolumn{1}{c|}{$71.98 \pm 0.19$} & \multicolumn{1}{c|}{$39.85 \pm 0.14$} & \multicolumn{1}{c|}{$78.43 \pm 0.39$} & $65.70 \pm 0.15$ & \multicolumn{1}{c|}{4}     & \multicolumn{1}{c|}{$84.73 \pm 0.24$}                & \multicolumn{1}{c|}{$57.74 \pm 0.49$} & \multicolumn{1}{c|}{$88.25 \pm 0.46$} & $79.81 \pm 0.12$  \\ \hline
\multicolumn{1}{|c|}{5}     & \multicolumn{1}{c|}{$71.59 \pm 0.35$} & \multicolumn{1}{c|}{$34.63 \pm 0.25$} & \multicolumn{1}{c|}{$79.61 \pm 0.57$} & $63.19 \pm 0.15$ & \multicolumn{1}{c|}{5}     & \multicolumn{1}{c|}{$84.09 \pm 0.14$} & \multicolumn{1}{c|}{$53.85 \pm 2.37$} & \multicolumn{1}{c|}{$88.85 \pm 0.56$} & $78.59 \pm 0.63$ \\ \hline
\multicolumn{5}{|c|}{Company C}                                                                                                                                        & \multicolumn{5}{c|}{Company D}                                                                                                                                                       \\ \hline
\multicolumn{1}{|c|}{Ratio} & \multicolumn{1}{c|}{Accuracy}         & \multicolumn{1}{c|}{Precision}        & \multicolumn{1}{c|}{Recall}           & $F_2$            & \multicolumn{1}{c|}{Ratio} & \multicolumn{1}{c|}{Accuracy}                        & \multicolumn{1}{c|}{Precision}        & \multicolumn{1}{c|}{Recall}           & $F_2$            \\ \hline
\multicolumn{1}{|c|}{3}     & \multicolumn{1}{c|}{$72.84 \pm 0.32$} & \multicolumn{1}{c|}{$47.32 \pm 0.38$} & \multicolumn{1}{c|}{$76.52 \pm 0.26$} & $68.11 \pm 0.12$ & \multicolumn{1}{c|}{3}     & \multicolumn{1}{c|}{$89.75 \pm 0.1$}                & \multicolumn{1}{c|}{$73.49 \pm 0.28$} & \multicolumn{1}{c|}{$92.1 \pm 0.17$} & $87.66 \pm 0.07$ \\ \hline
\multicolumn{1}{|c|}{4}     & \multicolumn{1}{c|}{$72.19 \pm 0.25$} & \multicolumn{1}{c|}{$39.99 \pm 0.23$} & \multicolumn{1}{c|}{$78.39 \pm 0.34$} & $65.76 \pm 0.15$ & \multicolumn{1}{c|}{4}     & \multicolumn{1}{c|}{$89.73 \pm 0.16$}                & \multicolumn{1}{c|}{$67.82 \pm 0.44$} & \multicolumn{1}{c|}{$92.62 \pm 0.19$} & $86.3 \pm 0.02$  \\ \hline
\multicolumn{1}{|c|}{5}     & \multicolumn{1}{c|}{$72.33 \pm 0.48$} & \multicolumn{1}{c|}{$35.29 \pm 0.04$} & \multicolumn{1}{c|}{$79.36 \pm 0.31$} & $63.5 \pm 0.13$  & \multicolumn{1}{c|}{5}     & \multicolumn{1}{c|}{$89.7 \pm 0.11$}                 & \multicolumn{1}{c|}{$62.87 \pm 0.33$} & \multicolumn{1}{c|}{$93.4 \pm 0.29$} & $85.13 \pm 0.11$  \\ \hline
\end{tabular}
\caption{In-depth comparison of candidate ratios of these four companies for training set}
\label{tab:comarison_candidate_train}
\end{table*}

\subsection{Random Forest Comparison}
\label{sub:compare_RF}

Table~\ref{tab:RF_vs_Our} shows a comparison of precision, recall, and $F_2$ score for our proposed deep learning framework (denoted as DL-AE) and random forest (denoted as RF) for four different companies, i.e., companies $A$, $B$, $C$, and $D$. Our framework outperforms random forest in terms of recall and $F_2$ score for all four companies, while random forest provides better precision. As we prioritize recall over precision, our framework is considered more suitable for our use case. The performance of both frameworks varies based on the company, which is dependent on the product and its audience. These results coupled with the conversion rate analysis demonstrate that our proposed deep learning framework outperforms random forest in real-world scenarios.

\begin{table}[h]
    \centering
    \begin{tabular}{|c|c|c|c|c|c|} \hline
        Company & Method & Accuracy & Precision & Recall & $F_2$\\ [0.5ex] \hline \hline 
        %$A$ & RF  & $79.48$ & $57.51$ & $63.27$ & $62.03$ \\ \hline
        $A$ & RF  & $79.53$ & $47.97$ & $47.44$ & $47.54$ \\ \hline
        $A$ & DL-AE & $70.82$ & $38.29$ & $76.95$ & $64.02$ \\ [0.5ex] \hline \hline

        $B$ & RF  & $87.3$ & $68.13$ & $70.16$ & $69.75$  \\ \hline
        $B$ & DL-AE & $83.96$ & $56.35$ & $85.37$ & $77.4$ \\ [0.5ex] \hline \hline

        $C$ & RF  & $79.37$ & $48.3$ & $46.06$ & $46.49$ \\ \hline
        $C$ & DL-AE & $71.08$ & $39.11$ & $74.12$ & $62.86$ \\ [0.5ex] \hline \hline

        $D$ & RF  & $90.77$ & $74.22$ & $81.91$ & $80.24$ \\ \hline
        $D$ & DL-AE & $88.41$ & $64.74$ & $90.1$ & $83.55$ \\ [0.5ex] \hline \hline

    \end{tabular}
    \caption{Comparison of random forest (RF) versus our proposed framework (DL-AE) for test test}
    \label{tab:RF_vs_Our}
\end{table}

Since the focus of this paper is on proposing the deep learning method for tabular data, we do not provide the details of how we tune our random forest framework. In this paper, we only provide the parameters of our random forest model; however, similar to deep learning, we have done rigorous testing in order to find the best parameters for our random forest framework. The depth of our random forest model ranges from  9-18, with 200-500 trees, and a minimum between 5  - .01 percent (training data size)  instances per node.

\subsection{Real-World Performance}
\label{sub:real-world}

In this section, we evaluate our proposed framework in the real world. We aim to see how our proposed framework works for prospecting via direct mail. In a prospecting campaign, the specific number of mail pieces that are sent to different households is determined by the company's budget and referred to as the campaign's \textbf{reach}. The number of targeted households that end up purchasing a product from that company within a specified time window (attribution window) are considered \textbf{converters}. Each individual transaction from one of these converters is considered a \textbf{conversion}. To calculate the performance for a prospecting campaign, we are mainly interested in the \textit{conversion rate} (denoted by \textit{CVR}), or the number of conversions (denoted by \textit{\#CNV}) divided by the reach. 

The output of our proposed framework gives a probability for each sample showing the likelihood of that sample being a prospecting customer. Based on these probabilities, we can rank the whole population and distribute reach according to this ranking. In other words, the ranking shows the priority that a household would have during a campaign mailing.

We evaluated the real-world performance of two companies, i.e., Company $E$ and Company $F$. These companies were selected as they are distinct from those used in the development and testing of our framework. This serves to demonstrate the generality of our framework and its ability to be applied to different products and audiences. Moreover, Company $E$ operates in the home services sector, while Company $F$ is in the home decoration sector.

In our Evaluation, we have tested different audiences within each company for both our proposed deep learning framework, which is denoted by \textit{DL-AE}, and random forest, denoted by \textit{RF} in Table~\ref{tab:real_worls_metric}. Testing different audiences with different sizes for a company will lead us to find new potential customers since each audience consists of a special sector of the company's customers.

\begin{table}[h]
    \centering
    \begin{tabular}{|c|c|c|c|c|c|} \hline
        Company & Method & Audience & Reach & \#CNV & CVR\\ [0.5ex] \hline \hline 
        $E$ & DL-AE  & $E1$ & $309,963$ & $1,043$ & $0.34\%$ \\ \hline
        $E$ & RF  & $E1$ & $154,981$ & $399$ & $0.26\%$ \\ [0.5ex] \hline \hline

        $E$ & DL-AE  & $E2$ & $101,109$ & $220$ & $0.22\%$ \\ \hline
        $E$ & RF  & $E2$ & $404,433$ & $592$ & $0.15\%$\\ [0.5ex] \hline \hline
        
        $F$ & DL-AE  & $F1$ & 119,076 & 3,679 & 3.09\% \\ \hline
        $F$ & RF & $F1$ & 77,385 & 2,245 & 2.90\% \\ [0.5ex] \hline \hline

        $F$ & DL-AE  & $F2$ & 61,881 & 2,003 & 3.24\% \\ \hline
        $F$ & RF & $F2$ & 91,658 & 2,771 & 3.02\% \\ [0.5ex] \hline \hline

    \end{tabular}
    \caption{Comparison of random forest (RF) versus our proposed framework (DL-AE) in real-world}
    \label{tab:real_worls_metric}
\end{table}

In Table~\ref{tab:real_worls_metric}, the size of audience $E1$ and $E2$ are $329,110$ and $390,273$ respectively. For company $F$, audience $F1$ and $F2$ size are $446,606$ and $347,179$. For privacy reasons for these companies, we do not reveal the difference between these audiences except for their size. 

As we can see in Table~\ref{tab:real_worls_metric}, the real-world performance of our proposed framework is better than the random forest. 

In addition to considering the conversion rate of a couple of campaigns for some companies, we consider the performance of company $E$ for multiple campaigns over the course of 6 months and compare how often our proposed architecture performs better than RF in practice. Over the course of 6 months, we have run 46 campaigns for this company, and the reach ranges from 20,000 to 500,00 based on the goal of the campaign and budget for that specific campaign. When evaluating the performance of 46 different campaigns using conversion rate, our proposed DL framework outperformed the random forest model 30 times and matched random forest performance 9 times. In production, the random forest model only performed better than our proposed DL framework 7 times. The above analysis corroborates the consistent performance of our proposed DL approach over time compared to the random forest, which is known as the best choice for tabular data.

\section{Discussion and Limitations}
\label{sec:discussion}

In this section, we discuss some implications of our work and some limitations.

\subsection{Label and Ground Truth}
\label{sub:truth}

To address the issue of prospecting, we selected our label $0$ by randomly sampling from the population excluding current customers. While some individuals in this sample may have never been exposed to the product, making it uncertain if they are truly label $0$, we chose this method as it is infeasible to determine with certainty whether an individual will never purchase the product. There is a possibility that they may become customers in the future. The random sample from across the entire United States allows us to conceptually learn how our client's customers compare to the average American. 

Additionally, testing our framework in the real world and calculating conversions does not necessarily indicate that non-converters are not customers. It only shows that they didn't choose to purchase the product during the current attribution window. Factors such as a different attribution window, increased ad exposure, altered messaging, or alternative offers may influence their decision to convert in the future.

It is important to note that there is no standard conversion rate that is considered acceptable. Conversion rates vary depending on a company's marketing goals, which can change from company to company and season to season. Whether the conversion rate of a specific campaign is reasonable depends on the company's objectives and current market conditions. 

\subsection{Interpretability}
\label{sub:interpretable}

Given the uncertainty surrounding label $0$, we evaluate the interpretability of our model using Shapley values~\cite{lundberg2017unified}. We calculate the impact of individual features on the prediction for a set of samples ranked at the top of our model's predictions. This analysis demonstrates that features related to the product have the greatest impact on the prediction. In other words, we aim to confirm that our model is functioning correctly and picking up relevant product-related features. Our analysis suggests that the uncertainty regarding label $0$ does not negatively impact the accuracy of our model. However, due to proprietary data restrictions from our data providers, we cannot reveal the specific features or products involved. We only mention this investigation to ensure that our model is functioning correctly.

\subsection{Privacy}
\label{sub:privacy}

During our model development and evaluation, we utilized a set of features from a single data provider source. To maintain privacy and comply with regulations, we have taken the following precautions:
\begin{enumerate}
    \item We only used data from large, publicly traded commercial data providers with established privacy and compliance processes. We did not combine this data with any other publicly accessible sources.

    \item  All data used in our model development and evaluation was anonymized.

    \item We did not use any sensitive personal information such as health records, credit card information, race, etc.

    \item We are in compliance with regulations outlined in the California Consumer Privacy Act (CCPA)~\cite{ccpa2023} which gives consumers the right to know, delete, opt-out, and the right to non-discrimination regarding the information collected by a business.
\end{enumerate}

\section{Conclusion}
\label{sec:conclusion}

This paper presents a novel deep learning framework for handling imbalanced tabular data in an applied real world scenario. The framework consists of two components: an autoencoder and a feed-forward neural network, which are designed to efficiently handle large datasets with numerous features. The performance of the proposed framework is evaluated through a real-world case study of direct mail prospecting advertisement. The study investigates important architecture selections such as the encoder size, feed-forward neural network architecture, and ratio, and compares the performance of the proposed model to that of a tree-based random forest model. The results show that the proposed framework outperforms the random forest in traditional metrics such as precision and recall, as well as in real-world performance in the prospecting campaign.

The proposed framework is general in nature and could be applied to other binary classification tasks. The authors intend to further explore this potential in future work by evaluating the framework on different tasks with varying feature sizes and numbers, and comparing its performance to tree-based models like random forest and XGBoost.

%\begin{acks}
%To Robert, for the bagels and explaining CMYK and color spaces.
%\end{acks}

\bibliographystyle{ACM-Reference-Format}
\bibliography{main}
\appendix
\section{Ratio Selections} 
\label{app:archs}

\begin{table}[]
    \centering
    \begin{tabular}{|c|c|c|c|c|c|} \hline
        Ratio & Train-Test & Accuracy & Precision & Recall & $F_2$\\ [0.5ex] \hline \hline 
        $1$ & Train & $0.731$ & $0.723$ & $0.75$ & $0.744$ \\ \hline
        $1$ & Test & $0.722$ & $0.722$ & $0.721$ & $0.722$ \\ [0.5ex] \hline \hline

        $2$ & Train & $0.725$ & $0.565$ & $0.769$ & $0.717$ \\ \hline
        $2$ & Test & $0.719$ & $0.558$ & $0.751$ & $0.702$ \\ [0.5ex] \hline \hline

        $3$ & Train & $0.723$ & $0.467$ & $0.778$ & $0.687$ \\ \hline
        $3$ & Test & $0.715$ & $0.459$ & $0.743$ & $0.661$ \\ [0.5ex] \hline \hline

        $4$ & Train & $0.718$ & $0.397$ & $0.789$ & $0.659$ \\ \hline
        $4$ & Test & $0.708$ & $0.383$ & $0.769$ & $0.640$ \\ [0.5ex] \hline \hline

        $5$ & Train & $0.711$ & $0.343$ & $0.804$ & $0.634$ \\ \hline
        $5$ & Test & $0.714$ & $0.34$ & $0.753$ & $0.606$ \\ [0.5ex] \hline \hline

        $6$ & Train & $0.712$ & $0.307$ & $0.807$ & $0.609$ \\ \hline
        $6$ & Test & $0.7$ & $0.292$ & $0.779$ & $0.584$ \\ [0.5ex] \hline \hline

        $7$ & Train & $0.711$ & $0.277$ & $0.816$ & $0.588$ \\ \hline
        $7$ & Test & $0.695$ & $0.26$ & $0.788$ & $0.56$ \\ [0.5ex] \hline \hline

        $8$ & Train & $0.708$ & $0.252$ & $0.829$ & $0.569$ \\ \hline
        $8$ & Test & $0.693$ & $0.234$ & $0.787$ & $0.535$ \\ [0.5ex] \hline \hline

        $9$ & Train & $0.709$ & $0.233$ & $0.836$ & $0.551$ \\ \hline
        $9$ & Test & $0.704$ & $0.221$ & $0.78$ & $0.518$ \\ [0.5ex] \hline \hline

        $10$ & Train & $0.705$ & $0.216$ & $0.849$ & $0.535$ \\ \hline
        $10$ & Test & $0.692$ & $0.197$ & $0.782$ & $0.491$ \\ \hline
    \end{tabular}
    \caption{Comparison of different ratio for company $A$}
    \label{tab:company_A_ratio}
\end{table}

\begin{table}[]
    \centering
    \begin{tabular}{|c|c|c|c|c|c|} \hline
        Ratio & Train-Test & Accuracy & Precision & Recall & $F_2$\\ [0.5ex] \hline \hline 
        $1$ & Train & $0.848$ & $0.849$ & $0.845$ & $0.846$ \\ \hline
        $1$ & Test & $0.844$ & $0.852$ & $0.834$ & $0.838$ \\ [0.5ex] \hline \hline

        $2$ & Train & $0.851$ & $0.736$ & $0.861$ & $0.833$ \\ \hline
        $2$ & Test & $0.842$ & $0.722$ & $0.848$ & $0.819$ \\ [0.5ex] \hline \hline

        $3$ & Train & $0.848$ & $0.645$ & $0.873$ & $0.816$ \\ \hline
        $3$ & Test & $0.833$ & $0.619$ & $0.849$ & $0.79$ \\ [0.5ex] \hline \hline

        $4$ & Train & $0.843$ & $0.569$ & $0.890$ & $0.800$ \\ \hline
        $4$ & Test & $0.828$ & $0.544$ & $0.869$ & $0.775$ \\ [0.5ex] \hline \hline

        $5$ & Train & $0.846$  & $0.524$ & $0.0.893$ & $0.783$ \\ \hline
        $5$ & Test & $0.835$ & $0.492$ & $0.863$ & $0.750$ \\ [0.5ex] \hline \hline

        $6$ & Train & $0.851$ & $0.488$ & $0.896$ & $0.767$ \\ \hline
        $6$ & Test & $0.842$ & $0.474$ & $0.85$ & $0.733$ \\ [0.5ex] \hline \hline

        $7$ & Train & $0.851$ & $0.451$ & $0.907$ & $0.754$ \\ \hline
        $7$ & Test & $0.827$ & $0.412$ & $0.867$ & $0.710$ \\ [0.5ex] \hline \hline

        $8$ & Train & $0.852$  & $0.423$ & $0.910$ & $0.74$ \\ \hline
        $8$ & Test & $0.845$ & $0.411$ & $0.851$ & $0.701$ \\ [0.5ex] \hline \hline

        $9$ & Train & $0.852$ & $0.396$ & $0.923$ & $0.729$ \\ \hline
        $9$ & Test & $0.84$ & $0.372$ & $0.854$ & $0.679$ \\ [0.5ex] \hline \hline
        
        $10$ & Train & $0.85$ & $0.371$ & $0.937$ & $0.718$ \\ \hline
        $10$ & Test & $0.835$ & $0.0.341$ & $0.864$ & $0.661$ \\ \hline
    \end{tabular}
    \caption{Comparison of different ratio for company $B$}
    \label{tab:company_B_ratio}
\end{table}

\begin{table}[]
    \centering
    \begin{tabular}{|c|c|c|c|c|c|} \hline
        Ratio & Train-Test & Accuracy & Precision & Recall & $F_2$\\ [0.5ex] \hline \hline 
        $1$ & Train & $0.729$  & $0.734$ & $0.716$ & $0.719$ \\ \hline
        $1$ & Test & $0.712$ & $0.723$ & $0.706$ & $0.709$ \\ [0.5ex] \hline \hline

        $2$ & Train & $0.731$ & $0.574$ & $0.748$ & $0.705$ \\ \hline
        $2$ & Test & $0.714$ & $0.563$ & $0.733$ & $0.691$ \\ [0.5ex] \hline \hline

        $3$ & Train & $0.724$ & $0.468$ & $0.769$ & $0.681$ \\ \hline
        $3$ & Test & $0.713$ & $0.459$ & $0.736$ & $0.657$ \\ [0.5ex] \hline \hline

        $4$ & Train & $0.719$ & $0.398$ & $0.789$ & $0.66$ \\ \hline
        $4$ & Test & $0.711$ & $0.389$ & $0.758$ & $0.637$ \\ [0.5ex] \hline \hline

        $5$ & Train & $0.723$ & $0.353$ & $0.793$ & $0.635$ \\ \hline
        $5$ & Test & $0.705$ & $0.326$ & $0.739$ & $0.59$ \\ [0.5ex] \hline \hline

        $6$ & Train & $0.717$ & $0.313$ & $0.817$ & $0.618$ \\ \hline
        $6$ & Test & $0.701$  & $0.287$ & $0.751$ & $0.567$ \\ [0.5ex] \hline \hline

        $7$ & Train & $0.732$ & $0.293$ & $0.81$ & $0.598$ \\ \hline
        $7$ & Test & $0.716$  & $0.267$ & $0.717$ & $0.536$ \\ [0.5ex] \hline \hline

        $8$ & Train & $0.726$ & $0.267$ & $0.841$ & $0.588$ \\ \hline
        $8$ & Test & $0.715$ & $0.241$ & $0.733$ & $0.521$ \\ [0.5ex] \hline \hline

        $9$ & Train & $0.721$ & $0.246$ & $0.866$ & $0.575$ \\ \hline
        $9$ & Test & $0.698$ & $0.213$ & $0.733$ & $0.492$ \\ [0.5ex] \hline \hline

        $10$ & Train & $0.729$ & $0.233$ & $0.866$ & $0.561$ \\ \hline
        $10$ & Test & $0.702$ & $0.198$ & $0.755$ & $0.483$ \\ \hline
    \end{tabular}
    \caption{Comparison of different ratio for company $C$}
    \label{tab:company_C_ratio}
\end{table}

\begin{table}[]
    \centering
    \begin{tabular}{|c|c|c|c|c|c|} \hline
        Ratio & Train-Test & Accuracy & Precision & Recall & $F_2$\\ [0.5ex] \hline \hline 
        $1$ & Train & $0.898$ & $0.895$ & $0.901$ & $0.90$ \\ \hline
        $1$ & Test & $0.885$ & $0.892$ & $0.877$ & $0.88$ \\ [0.5ex] \hline \hline

        $2$ & Train & $0.898$ & $0.806$ & $0.912$ & $0.889$ \\ \hline
        $2$ & Test & $0.892$  & $0.801$ & $0.899$ & $0.878$ \\ [0.5ex] \hline \hline

        $3$ & Train & $0.897$ & $0.733$ & $0.923$ & $0.878$ \\ \hline
        $3$ & Test & $0.892$ & $0.737$ & $0.904$ & $0.865$ \\ [0.5ex] \hline \hline

        $4$ & Train & $0.899$ & $0.684$ & $0.923$ & $0.863$ \\ \hline
        $4$ & Test & $0.884$ & $0.647$ & $0.901$ & $0.836$ \\ [0.5ex] \hline \hline

        $5$ & Train & $0.898$ & $0.630$ & $0.935$ & $0.852$ \\ \hline
        $5$ & Test & $0.881$ & $0.594$ & $0.898$ & $0.815$ \\ [0.5ex] \hline \hline

        $6$ & Train & $0.896$ & $0.584$ & $0.944$ & $0.840$ \\ \hline
        $6$ & Test & $0.888$ & $0.566$ & $0.900$ & $0.805$ \\ [0.5ex] \hline \hline

        $7$ & Train & $0.894$ & $0.544$ & $0.953$ & $0.829$ \\ \hline
        $7$ & Test & $0.884$ & $0.513$ & $0.905$ & $0.785$ \\ [0.5ex] \hline \hline

        $8$ & Train & $0.899$ & $0.526$ & $0.954$ & $0.820$ \\ \hline
        $8$ & Test & $0.887$ & $0.491$ & $0.9$ & $0.771$ \\ [0.5ex] \hline \hline

        $9$ & Train & $0.899$ & $0.497$ & $0.961$ & $0.810$ \\ \hline
        $9$ & Test & $0.9$ & $0.472$ & $0.903$ & $0.763$ \\ [0.5ex] \hline \hline

        $10$ & Train & $0.902$ & $0.48$ & $0.967$ & $0.804$ \\ \hline
        $10$ & Test & $0.891$ & $0.452$ & $0.895$ & $0.748$ \\ \hline
    \end{tabular}
    \caption{Comparison of different ratio for company $D$}
    \label{tab:company_D_ratio}
\end{table}

\end{document}